\documentclass[sigconf]{acmart}

\AtBeginDocument{%
  \providecommand\BibTeX{{%
    \normalfont B\kern-0.5em{\scshape i\kern-0.25em b}\kern-0.8em\TeX}}}

\copyrightyear{2023}
\acmYear{2023}
\setcopyright{acmlicensed}\acmConference[SC-W 2023]{Workshops of The
International Conference on High Performance Computing, Network, Storage,
and Analysis}{November 12--17, 2023}{Denver, CO, USA}
\acmBooktitle{Workshops of The International Conference on High Performance
Computing, Network, Storage, and Analysis (SC-W 2023), November 12--17,
2023, Denver, CO, USA}
\acmPrice{15.00}
\acmDOI{10.1145/3624062.3624614}
\acmISBN{979-8-4007-0785-8/23/11}

\newcommand\ottwo{\texttt{ot2}}
\newcommand\pffour{\texttt{pf400}}
\newcommand\sciclops{\texttt{sciclops}}
\newcommand\camera{\texttt{camera}}

\newcommand\barty{\texttt{barty}}

\newcommand\wei{WEI}

\title{Exploring Benchmarks for Self-Driving Labs using Color Matching}

\author{Tobias Ginsburg}
\affiliation{%
   \institution{Argonne National Laboratory}
  \streetaddress{9600 S. Cass Ave}
  \city{Lemont}
  \country{United States}}
\email{tginsburg@anl.gov}

\author{Kyle Hippe}
\affiliation{%
  \institution{Argonne National Laboratory}
  \streetaddress{9600 S. Cass Ave}
  \city{Lemont}
  \country{United States}}
\email{khippe@anl.gov}

\author{Ryan Lewis}
\affiliation{%
  \institution{Argonne National Laboratory}
  \streetaddress{9600 S. Cass Ave}
  \city{Lemont}
  \country{United States}}
\email{ryan.lewis@anl.gov}

\author{Doga Ozgulbas}
\affiliation{%
  \institution{Argonne National Laboratory}
  \streetaddress{9600 S. Cass Ave}
  \city{Lemont}
  \country{United States}}
\email{dozgulbas@anl.gov}

\author{Aileen Cleary}
\affiliation{%
  \institution{Northwestern University}
  \streetaddress{633 Clark Street}
  \city{Evanston}
  \country{United States}}
\email{aileencleary2026@u.northwestern.edu}

\author{Rory Butler}
\affiliation{%
  \institution{UChicago \& Argonne National Lab}
  \streetaddress{9600 S. Cass Ave}
  \city{Lemont}
  \country{United States}}
\email{rorymb@uchicago.edu}

\author{Casey Stone}
\affiliation{%
  \institution{Argonne National Laboratory}
  \streetaddress{9600 S. Cass Ave}
  \city{Lemont}
  \country{United States}}
\email{cstone@anl.gov}

\author{Abraham Stroka}
\affiliation{%
  \institution{Argonne National Laboratory}
  \streetaddress{9600 S. Cass Ave}
  \city{Lemont}
  \country{United States}}
\email{astroka@anl.gov}

\author{Rafael Vescovi}
\affiliation{%
  \institution{Argonne National Laboratory}
  \streetaddress{9600 S. Cass Ave}
  \city{Lemont}
  \country{United States}}
\email{ravescovi@anl.gov}

\author{Ian Foster}
\affiliation{%
  \institution{UChicago \& Argonne National Lab}
  \streetaddress{9600 S. Cass Ave}
  \city{Lemont}
  \country{United States}}
\email{foster@anl.gov}

\begin{document}

\renewcommand{\shortauthors}{Ginsburg et al.}

\begin{abstract}
Self Driving Labs (SDLs) that combine automation of experimental procedures with autonomous decision making are gaining popularity as a means of increasing the throughput of scientific workflows. The task of identifying quantities of supplied colored pigments that match a target color, the \textit{color matching problem}, provides a simple and flexible SDL test case, as it requires experiment proposal, sample creation, and sample analysis, three common components in autonomous discovery applications. 
We present a robotic solution to the color matching problem that allows for fully autonomous execution of a color matching protocol.
Our solution leverages the WEI science factory platform to enable portability across different robotic hardware, the use of alternative optimization  methods for continuous refinement, and automated publication of results for experiment tracking and post-hoc analysis.
%We present a modular, easily retargetable robotic solution to the color matching problem that allows for fully autonomous execution of a color matching protocol, with pluggable optimization approaches allowing for continuous refinement and automated publication of results facilitating experiment tracking and post-hoc analysis. 
\end{abstract}
%% TODO: Update these?
\keywords{Automation, Self-Driving Labs, Color Matching, Artificial Intelligence}
\maketitle
% \section*{Questions\\Tasks}
% \begin{itemize}
%     \item{Add names}
%     \item{review intro}
%     % \item{Add Wireframe}
%     % \item{Review min workcell}
%     % \item{Add models to the instruments used and reference}
  
%     % \item{CHE/REF models}
%     % \item{Solver (kyle)}
%     % \item{Color Analysis (Rory) - Did in subsection "Image Processing"}
%     % \item{Ot2 (Abe + Kyle)}
   
%     % \item{globus portal + acknowledgements}
% \end{itemize}

\begin{figure*}[htbp]
    \centering
    \includegraphics[width=\textwidth]{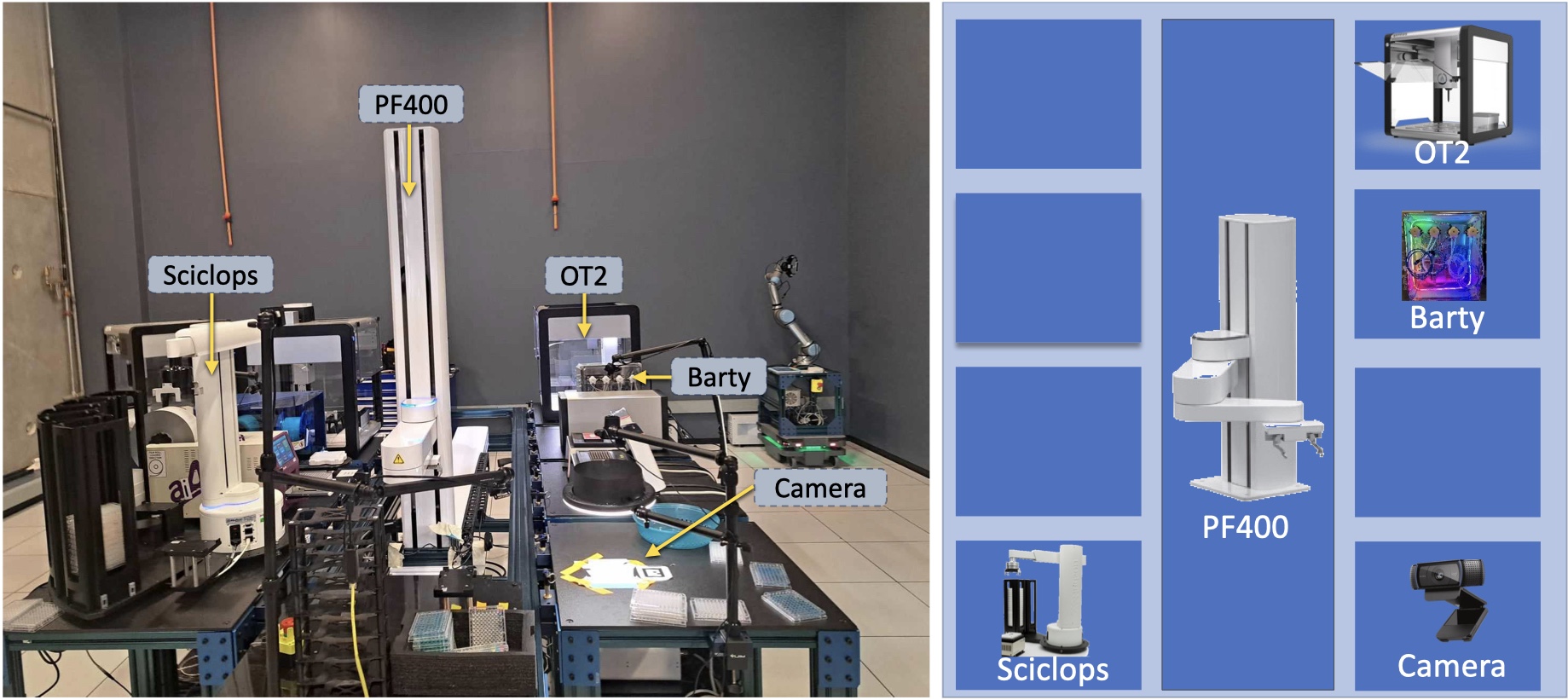}

    \caption{A photograph and diagram of the robotic workcell used for the color picking experiment. The \sciclops{} picks up a 96-well plate from its plate storage towers, and transfers it to its exchange location. The \pffour{} then transfers the plate to the \ottwo{}, which mixes the three target colors. When the liquid reservoirs in the system are empty, the custom robot, \barty{}, refills them by using peristaltic pumps. Once the mixing is completed, the plate is transferred to the Camera location to be imaged. The plate is then looped between the camera and the OT2 until the experiment is over.}
    \label{fig:wc}
\end{figure*}

\section{Introduction}
% Alternative introduction, feel free to harvest what you would like :) 
Self-Driving Laboratories (SDLs) combine automation of individual experimental steps with automated decision procedures (e.g., optimization or AI methods) for selecting new experiments.
By thus enabling closed-loop, autonomous discovery, SDLs can accelerate scientific discovery by enhancing precision, efficiency, and repeatability of experimental protocols~\cite{hase2019next,macleod2020self,vriza2023self,abolhasani2023rise,stach2021autonomous}. 
When combined with programming frameworks~\cite{steiner2019organic,vrana2021aquarium,roch2020chemos,scifac} that facilitate the substitution of alternative experimental apparatus, experimental protocols, and decision procedures, SDLs become broadly applicable, multi-purpose research instruments.

The color-mixing problem~\cite{roch2020chemos,baird2023building} has been employed by several SDL groups for pedagogical purposes and to permit low-cost experimentation with alternative decision procedures.
The task here is to identify quantities of a small number of provided colored pigments that, when mixed, best match a specified target color~\cite{roch2020chemos,baird2023building}.
While simple, this problem captures important elements of SDL applications: in particular, the need to propose samples, create samples, characterize samples, and process characterization results to generate new proposals.

The authors and their colleagues in Argonne's Rapid Prototyping Lab (RPL) have created a modular, re-configurable robotic workcell to enable multiple robotic workflows on a shared set of hardware. 
Using this system, we have implemented a color mixing application, \textit{color picker}, that employs five devices that serve other roles in other workflows. The color mixing is performed fully autonomously, with everything from fetching a new plate to analyzing the color image to uploading data to the web performed completely without human intervention. Over time, we were able to increase the lab's reliability while measuring its performance across a number of relevant metrics, showing the color mixing application's usefulness as an SDL benchmark. 

An earlier version of the color picker application was described in a previous paper~\cite{scifac}. Here we describe extensions to that work that incorporate an additional device, the Barty liquid replenisher, provide a more detailed description of the application, and discuss its use as a possible SDL benchmark.

\section{Methods}

We describe in turn the color-picker problem, our workcell architecture, our implementation of the problem on the workcell, and other aspects of our approach.

\subsection{Application}

The color picker application that we present here mixes four component liquids, specifically cyan, yellow, magenta, and black dyes, in a microplate to attempt to produce a liquid sample that matches a target color as closely as possible. 

This process consists of three steps. First, our optimization algorithm leverages its (initially empty) set of data obtained to date to propose a set of experiments to perform, expressed as a set of volumes for each liquid. Second, our robotic workcell is instructed to produce these samples by dispensing and mixing the appropriate quantities of each liquid.
Third, the workcell is instructed to measure the colors of the samples produced.
The results from the third step are then
fed back to step~1, until a termination criteria is satisfied. We accomplish these various tasks by using the execution framework described by Vescovi et al.~\cite{scifac}.

\begin{figure*}[htbp]
    \centering
    \includegraphics[width=\textwidth]{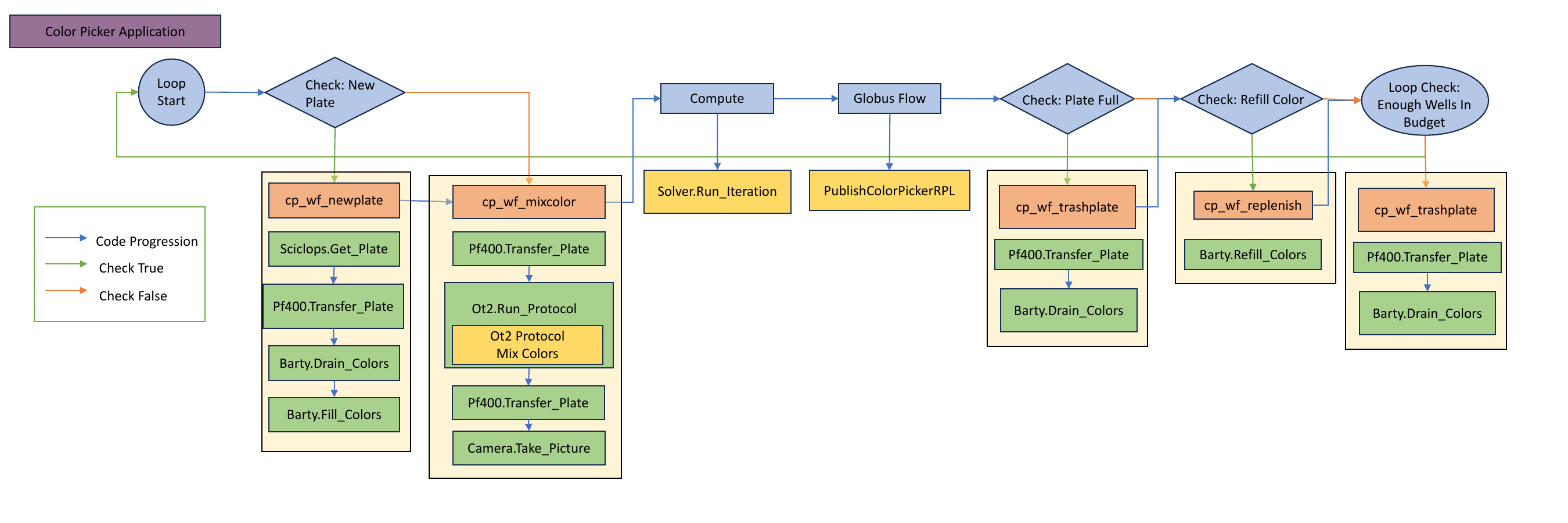}

\vspace{-2ex}
    
    \caption{The color-picker application. The Python code, \texttt{color\_picker\_app.py}, implements logic that  runs three distinct \wei{} flows, with the second (plus associated \texttt{publish} and \texttt{compute} steps) called repeatedly until termination criteria are satisfied.
    The orange box below the \ottwo{} \texttt{run\_protocol} action gives the name of the protocol file.
    Module names are as in Section~\ref{sec:modules}. %\autoref{tab:modules}.
    }
    \label{fig:cp_jpg}
\end{figure*}
\subsection{Workcell}\label{sec:modules}
% four separate instruments?

We have defined in previous work a modular architecture for SDLs in which \textit{modules} encapsulating scientific instruments and other devices (e.g., plate crane) are linked with manipulators to form \textit{workcells}~\cite{scifac}. 
Each \textit{module} is represented by a software abstraction that exposes a single device and, via interface methods, the actions that the device can perform; a declarative YAML notation is used to specify how a workcell is configured from a set of modules.

Users can specify, again using a declarative notation, \textit{workflows} that perform sets of actions on modules. 
They can also write Python \textit{applications} that run one or more workflows on specified workcell(s) and typically also invoke associated computational and data manipulation steps, for example to run solvers and publish results. 
Workflows can be reused in different applications, and modules in different workflows, and workflows can be retargeted to different modules and workcells that provide comparable capabilities. 
Workflow steps are translated into commands sent to computers connected to devices, 
%using a variety of protocols including REST-style HTTPS and ROS-node messaging.  These clients 
which then call driver functions specific to their attached device. 

The color picker application that we present here targets five modules: plate crane, manipulator, liquid handler, liquid replenisher, and camera.
In our experiments, we execute this application on the Argonne Rapid Prototyping Lab workcell, a configuration that comprises 10 modules useful for a variety of experimental protocols in biology and education. 
Here, we use just the five modules shown in \autoref{fig:wc}, and as specified in the YAML file at \url{https://bit.ly/3ZCVniT}.
%\url{https://github.com/AD-SDL/rpl_workcell/blob/main/rpl_modular_workcell.yaml}.

%A workcell is a collection of multiple autonomous devices, including measurement instruments, manipulators, and devices that perform specific actions, like sealing a plate. Each device is represented as a \textit{module}, a software abstraction that exposes a single device and, via interface methods, the actions that the devicce can perform. We show the five modules that we use for the color matching application in \autoref{fig:wc}. 

The complete RPL workcell also includes (not shown) modules used for PCR and for growing and analyzing cells, but this application targets only those depicted in \autoref{fig:wc}:
\begin{itemize}
    \item
    \sciclops{} is the Hudson SciClops Microplate Handler, a microplate storage and staging system that can access multiple storage towers, facilitating the housing of plates.
    \item
    \pffour{} is the workcell's manipulator, a robotic arm used to transfer microplates between different plate stations.  Operating on a rail mechanism, this robot acts as the central transportation unit within the workcell. Its core function is to shuttle microplates between different modules. 
    \item
    % four separate color reservoirs **
    \ottwo{} is an automatic pipetting device that contains four separate color reservoirs and a set of pipette tips. 
    Once the \pffour has delivered a plate to the \ottwo{} deck, it mixes liquids in the proportions set by the optimization algorithm to generate new sample colors. 
    \item
    \barty{} is a robot developed in RPL with four peristaltic pumps that transfer liquid from large storage vessels to the reservoirs of the the \ottwo{}.
    Our application instructs \barty{} to refill the \ottwo{} reservoirs periodically so that experiments can run for extended periods. 
    \item
    \texttt{camera} is a Logitech webcam mounted with a ring light that is used to capture images of the microplate. This module incorporates a microplate mount designed to allow the \pffour{} to place the microplate in the same location each time. 
\end{itemize}

\subsection{The Color Picker Application}

Our color picker application (accessible at 
\url{https://bit.ly/3ZCigTB})
%\url{https://github.com/AD-SDL/rpl_workcell/blob/main/color_picker_app/color_picker_application.py})
engages four workflows (see the folder \url{https://bit.ly/45bzFUb}),
%\url{https://github.com/AD-SDL/rpl_workcell/blob/main/color_picker_app/workflows}), 
as shown in \autoref{fig:cp_jpg}. 
It proceeds as follows:

%\vspace{-10mm}

\begin{enumerate}
\item 
    If a new plate is needed (as when starting or when the previous plate was full), call workflow \texttt{cp\_wf\_newplate} to instruct modules as follows:
    \begin{enumerate}
        \item
 \pffour{}: Retrieve new plate from \sciclops{}, place it at \texttt{camera}
    \item \barty{}: Fill \ottwo{} reservoirs
\end{enumerate}
\item 
    Run workflow \texttt{cp\_wf\_mix\_colors} to:
    \begin{enumerate}
    \item
   \pffour{}: Transfer the plate to \ottwo{}
    \item \ottwo{}: Dispense and combine specified pigment amounts
    \item \pffour{}: Return plate to  \texttt{camera} 
    \item \texttt{camera}: Photograph the plate.
       \end{enumerate}
\item 
    Process the image as described in Section~\ref{sec:ip}.
\item 
    Publish the obtained data, as described below.  
\item 
    Invoke a specified Color Picking Solver to evaluate the data and, if the termination criteria are not met, select the next set of colors.
\item If the plate is full, run workflow \texttt{cp\_wf\_trashplate} to dispose of the plate and drain the \ottwo{} reservoirs using \barty{}. 
%The \texttt{cp\_wf\_newplate}workflow is then run again to replace the microplate and reset the \ottwo{} reservoirs.
\item If the reservoirs need refilling, run workflow \texttt{cp\_wf\_replenish} to drain and refill them.
\end{enumerate}

Once termination criteria are satisfied (e.g., target color matched or resources exhausted), the application 
runs \texttt{cp\_wf\_trashplate} again to finalize the experiment. 

For each workflow that is run, a file is created that details the step names run, their start time, end time and total duration. These files are saved locally to the maching running the workflow manager.

\begin{figure*}
 \includegraphics[width=0.9\linewidth]{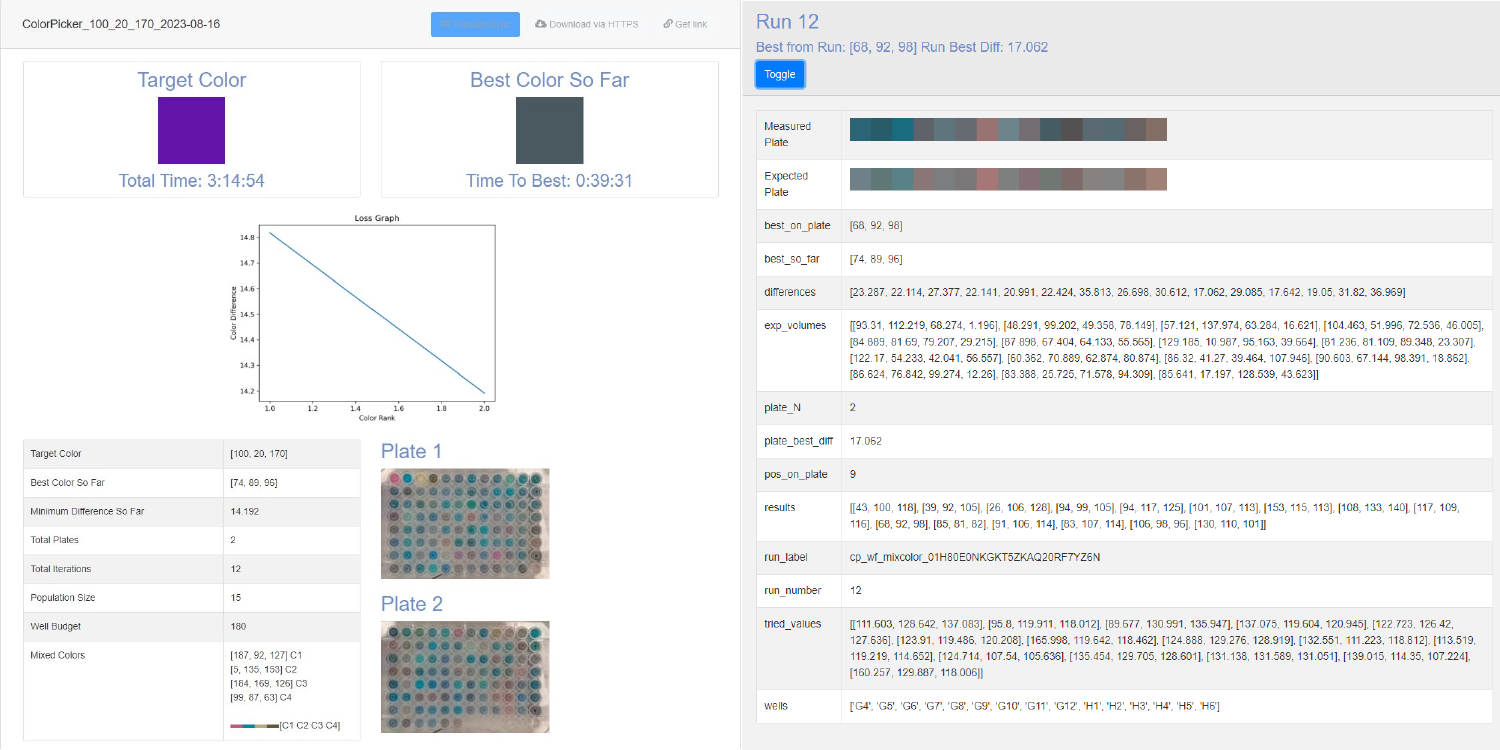}
 \caption{Two views of a Globus Search portal for data generated by the color-picker application, at \url{https://acdc.alcf.anl.gov}. \textit{Left}: Summary view for an experiment performed on August 16th, 2023, involving 12 runs each with 15 samples, for a total of 180 experiments. The images are those taken by the camera. 
    \textit{Right}: Detailed data from run \#12.}
    \label{fig:acdc}
\end{figure*}

The publication step engages a Globus flow 
\cite{chard2023globus}
to %uses the Gladier Publish tool to 
publish data to the ALCF Community Data Co-Op (ACDC) data portal (\autoref{fig:acdc}). For each run, the data created includes the colors produced, the timing of each step, the scoring results from the solver, and the raw plate images for quality control.

\subsection{Image Processing}
\label{sec:ip}

% Not sure where to put this section.
% https://github.com/AD-SDL/rpl_workcell/blob/main/color_picker_app/tools/plate_color_analysis.py

To process the webcam images for color detections, we station the plate at a known distance from an ArUco marker \cite{Aruco}. Using OpenCV \cite{opencv_library}, we detect the ArUco marker in the image, and use the size and position of the marker to determine the approximate pixel-coordinate boundaries of the microplate. To account for potential shifting in the camera position, we further refine the location of the microplate by drawing on a particular physical characteristic of the plate: the circular wells. With the HoughCircles algorithm from OpenCV, we can detect circular features in the image to precisely identify the center of wells. As this method is prone to false negatives, we further align a grid to all well-sized circles within the approximate plate position, and use this grid's size and orientation to predict the center points for all wells in the image, even those originally missed by the HoughCircles algorithm. Finally, the detected colors at the well center points are reported to the optimization algorithm.

\subsection{Color Picking Solvers}

The color picking problem admits to an analytic solution, given accurate models of how colors combine and the properties of our color sensor. However, treating the problem as a black box, as we do here, allows us to employ the problem as a surrogate for more complex problems and to experiment with different decision procedures in a simple setting.
We have implemented to date two such decision procedures, a simple evolutionary solver (a genetic algorithm) and a Bayesian solver, thus demonstrating the ability to run multiple optimization algorithms without changes to other elements of the system. 

%We present in Section~3 results obtained when using a simple evolutionary solver. 
%In order to validate our ability to run substitute 
%We have also implemented a number of other decision procedures, including the two described in the following.
%This solver allowed us to demonstrate the capacity for multiple different optimization algorithms to be run using the same system. 

%This allows us to evaluate the platform under workloads closer to those which it is expected to encounter in more complex applications.
%We have implemented multiple decision procedures for selecting experiments, so as to demonstrate the system's flexibility and ability to incorporate different optimization strategies. We describe two such methods here.

%Baird describes a variety of 

Genetic algorithms (GAs) are optimization and search heuristics inspired by natural evolution. Initially populated with random or heuristically chosen solutions, each iteration---termed a generation---evaluates individuals based on a fitness function. The fittest individuals are selected, and the remainder of the population is augmented. Over multiple generations, the population converges towards optimal or near-optimal solutions. For the initial population, points are sampled from a uniform grid of proper dimensions (corresponding to the number of mixing colors). Once a population of colors with grades (floating point values representing \textit{delta e} distance to the target) a new population is created. The most accurate element of the previous population is propagated into the new generation. One third of the new population is created by randomly selecting two elements of the previous population and taking the average of them. One third of the population is created by taking a random element of the previous population and randomly shifting its ratios. The final third of the population is created by randomly creating a new set of ratios.  The code for this is at \url{https://bit.ly/3rGxI4m}. %\url{https://github.com/AD-SDL/rpl_workcell/blob/main/color_picker_app/solvers/evolutionary_solver.py}.
We present results obtained with the genetic algorithm in Figure~4.

We also implemented a Bayesian optimization method based on scikit-learn~\cite{scikit-learn}: see \url{https://bit.ly/3EVDNgs}. Bayesian optimization leverages a surrogate probabilistic model, commonly Gaussian Processes, to approximate the objective function and iteratively refines this based on evaluations.
We do not present results obtained with this approach as they do not yield a systematic improvement over the genetic algorithm.
%\url{https://github.com/AD-SDL/rpl_workcell/blob/main/color_picker_app/solvers/bayes_solver.py}.

\begin{figure*}[htbp]
    \centering
    \includegraphics[width=0.85\textwidth]{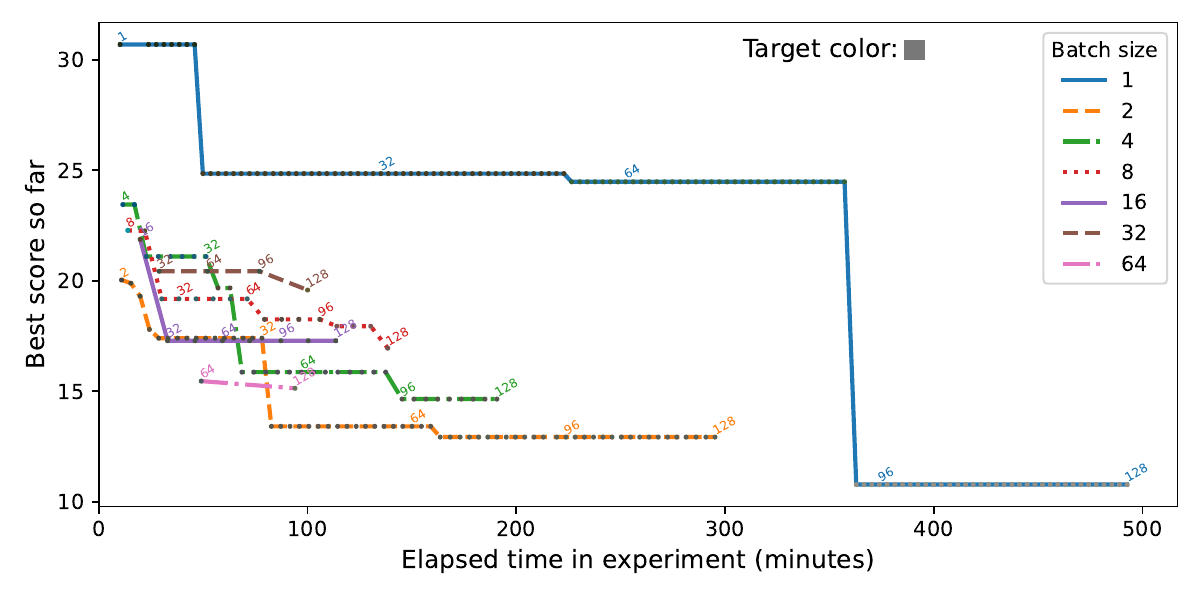}
    \caption{Results of seven experiments, in each of which the color picker application creates and evaluates 128 samples, in batches of an experiment-specific size 
    \emph{B} = 1, 2, 4, 8, 16, 32, and 64.
    In each experiment, the target color is RGB=(120,120,120), the first sample(s) are chosen at random, and later samples are chosen by applying a solver algorithm to \camera{} images.
    Each dot has as x-value the elapsed time in the experiment and as y-value the Euclidean distance in three-dimensional color space
    between the target color and the best color seen so far in the experiment. The evolutionary algorithm used has random elements, which means that improvement between iterations is not guaranteed, leading to the long flat stretches in the graph.
    The numbers in the graph represent selected sample sequence numbers. 
    Results depend %significantly 
    on the original random guesses, but overall, as we might expect, experiments with smaller batch sizes achieve lower scores, but take longer to run. (Figure adapted from \cite{scifac}.)
    }
    \label{fig:cp_data}
\end{figure*}

\section{Results}

We present results obtained when running the color picker application with the simple evolutionary solver, a fixed target color, and the total number of samples fixed at N=128. We varied the batch size B across different experiments by powers of two from 1 to 64. 

The results, as depicted in \autoref{fig:cp_data}, show outcomes from varying batch sizes. Each dot in the figure represents the time elapsed in the experiment (x-axis) against the Euclidean distance in the three-dimensional color space between the target color and the best-matched color at that point (y-axis).

It is evident from the results that the initial random guesses influence the outcomes. However, a general trend observed is that experiments with smaller batch sizes took longer to run but were more accurate in color matching.

The data publication capabilities allowed us to capture these results efficiently. \autoref{fig:acdc} provides two views of the Globus Search portal, detailing data generated by the application.

Our longest run with $B$=1 (i.e., a single sample per iteration) completes in 8 hours and 12 minutes without human intervention. This run consisted of 387 distinct robotic actions that were performed without error, and 128 distinct data upload steps that allowed for fine-grained tracking of the progress of the experiment. Data uploads occurred on average every 3 minutes and 48 seconds, and required moving the microplate from the camera module to the \ottwo{} and back, meaning the \pffour{} had to pick and place the microplate precisely twice per time period.

\section{Discussion}

We have presented a solution to the color matching problem as an exemplary implementation of experiment-in-the-loop computing.
Leveraging a recent proposed modular SDL architecture~\cite{scifac}, our implementation combines fine-grained control of various experimental apparatus with invocations of computing and data resources, accessed and coordinated by using Globus automation service mechanisms~\cite{vescovi2022linking}.
Continuous publishing of results permits monitoring of progress, and its modular architecture allows for the substitution of alternative instruments, optimization solvers, computing resources, and storage resources.
While certainly not extreme-scale, we expect this application to be of interest to XLOOP attendees as a test case for their own experiment-in-the-loop approaches.

The color-mixing application allows us to test aspects of a setup that we believe are necessary for any multipurpose self-driving lab application. Specifically, it requires a synthesis step and a data acquisition step, with distinct instruments for each step.  We suggest this sets a self-driving lab apart from single instrument stations that can perform multiple functions. It also requires multiple iterations of a similar process, necessary for using optimization or machine-learning algorithms to generate samples. The reliability with which we were able to run it allowed us to test the resiliency of our workcell infrastructure and learn more about the nature of errors that could occur, and also to compare different optimization algorithms. 
Because the color picking experiment uses inert reagents, it can be run on multiple different forms of hardware. As such, we believe it could serve as a benchmark application for self-driving labs. Specifically, we will discuss metrics around a color-mixing run with a batch size of 1, as this required the most robot coordination, and ran for the greatest amount of time. 

\begin{table}
    \caption{Proposed metrics for self-driving labs and our best results for a color picker batch size of 1.}
    \label{tab:sdl_metrics}
    \centering
    \begin{tabular}{|c|c|} \hline 
         \textbf{Metric}& \textbf{Value for B = 1}\\ \hline 
          Time without humans & 8 hours 12 mins\\ \hline 
          Completed commands without humans& 387\\ \hline 
         Synthesis time & 5 hours 10 mins\\ \hline 
         Transfer time& 3 hours 2 mins\\ \hline 
         Total colors mixed& 128\\ \hline 
 Time per color & 4 mins\\ \hline
    \end{tabular}
\end{table}

As the number of multi-purpose self-driving labs increases, the creation of metrics to compare and contrast their capabilities becomes relevant both to designers hoping to measure their performance of their systems and to researchers hoping to best utilize the available resources. We propose three metrics for evaluating the performance of the color-mixing experiment.  
We suggest that metrics such as these can provide a basis for comparison across different SDL implementations, and ultimately may allow facilities to convey their capabilities accurately to possible users. 

The first metric is \textit{time without human input} (TWH), equivalent to the longest time that an experiment ran without human intervention. This metric gives a sense of how much human time is being saved by automating the lab. 
However, it does not necessarily reflect a level of automation, as it is sensitive to how long individual actions take to run. 
Our best run according to this metric was 8 hours and 12 minutes. 

The second metric is \textit{commands completed without human input} (CCWH), i.e., the number of commands sent and successfully executed by the instruments over the course of the experiment without human intervention. A \textit{command} is defined as one or more actions carried out consecutively by a single instrument without input from the control system. A \textit{completed} command is one for which the instrument successfully performs each constituent action and reports success to the system. In our experience, most failures occur during reception and processing of commands, making CCWH a good measure of the resiliency of the SDL's communications and the reliability of its constituent instruments. Our best run according to this metric involved 387 successful commands. 
%We note that the TWH and CCWH can interact in interesting ways IAN STOPPING FOR A MOMENT
An interesting future experiment would involve integrating additional OT2s in our workflow, so that multiple plates of colors could be mixed at once
%An interesting study of the interaction of these two metrics would be the integration of another OT2 into our workflow, allowing for multiple plates of colors to be mixed at once. 
This would lead to an increase in CCWH, but potentially a lower TWH for the same experimental results.  

The third metric, which speaks to the efficiency of the lab overall, is \textit{time per color}, i.e., total experiment run time divided by the number of color samples produced.  This metric captures the per-sample efficiency and allows comparisons across different self-driving lab configurations and levels of parallelism. We achieved around four minutes per color for our B=1 run. We can also divide the total run time into \textit{synthesis time}, that used specifically to mix colors, and \textit{transfer time}, that used to move samples between instruments, which can help provide more information on where the bottlenecks in the system lie. Our B=1 run achieved 5 hours 10 minutes synthesis time and 3 hours 2 minutes transfer time, meaning that 63\% of the total time time was spent mixing colors using the OT2, possibly an area for improvement. 

%The color-mixing application has the advantage of involving no hazardous materials. If users do not want to expose their lab to colored reagents, for whatever reason, some other inert sample, such as a mixed water, could be used. 

In future work we would like to integrate our system with  Baird and Sparks' closed-loop spectroscopy lab code~\cite{baird2023building} so as to permit experimentation with their various optimization codes and different search approaches.
%While modifications would be necessary to incorporate publication and state tracking, this work would produce a standard benchmark for color mixing that would allow users to fit their own implementation into our broader data-collection framework. 
%In conclusion, the strides made in this study open doors for more extensive applications in the realm of autonomous robotic operations, extending far beyond just color matching.

%\textbf{Continuous operation}:
%Large-scale, long-term SDL operation requires the automation of support functions (e.g., replenishing consumables, disposing of waste, correcting operational errors) that in simpler settings might be handled by humans.
%We propose time-without-human-input as a useful metric for quantifying the level of automation achieved for both individual applications and a complete science factory running a mixed workload.

\begin{acks}
%We are grateful to Argonne colleagues with whom we have worked on SDLs, including Gyorgy Babnigg, Mike Papka,  Rick Stevens, and others. 
We thank Eric Codrea, Yuanjian Liu, 
%Priyanka Setty, 
and other students for their contributions, and Ryan Chard, Nickolaus Saint, and others in the Globus team for their ongoing support.
%We have benefited from conversations with many working in this area, including Sterling Baird, Andy Cooper, Lee Cronin, Jason Hattrick-Simpers, Ross King, Phil Maffettone, and Joshua Schreier. 
%This work would not have been possible without much appreciated support from the leadership and staff of Argonne's Leadership Computing Facility and Advanced Photon Source.
This work was supported in part by Laboratory Directed Research and Development funds at Argonne National Laboratory from the U.S.\ Department of Energy under Contract DE-AC02-06CH11357.

\end{acks}

\bibliographystyle{ACM-Reference-Format}
\bibliography{refs}

\end{document}